\documentclass[11pt]{article}

\usepackage{arxiv}

\usepackage[utf8]{inputenc}
\usepackage[T1]{fontenc}
\usepackage{lmodern}
\AtBeginDocument{\DeclareFontShape{T1}{lmr}{bx}{sc}{<->ssub*lmr/m/sc}{}}

\usepackage{amsmath}
\usepackage{amsfonts}
\usepackage{amssymb}

\usepackage{booktabs}       
\usepackage{array}           

\usepackage{graphicx}        
\usepackage{subcaption}      
\usepackage{float}           

\usepackage{microtype}       
\usepackage[colorlinks=true,linkcolor=linkblue,citecolor=linkblue,urlcolor=linkblue]{hyperref}
\usepackage{url}

\usepackage[numbers,sort&compress]{natbib}

\usepackage{xspace}          

\newcommand{\ours}{\textsc{Synthetic-Child}\xspace}   

\newcommand{\eg}{\textit{e.g.}\xspace}

\newcommand{\etal}{\textit{et al.}\xspace}

\title{%
  \ours: An AIGC-Based Synthetic Data Pipeline \\
  for Privacy-Preserving Child Posture Estimation%
}

\author{
  Taowen Zeng \\
  Independent Researcher \\
  \texttt{sharkfujin@gmail.com}
}

\date{}  

\begin{document}
\maketitle

\begin{abstract}
Accurate child posture estimation is critical for AI-powered study companion devices, yet collecting large-scale annotated datasets of children is both expensive and ethically prohibitive due to privacy concerns.
We present \ours, an AIGC-based synthetic data pipeline that produces photorealistic child posture training images with ground-truth-projected keypoint annotations, requiring \emph{zero} real child photographs.
The pipeline comprises four stages:
(1)~a programmable 3D child body model (SMPL-X) in Blender generates diverse desk-study poses with IK-constrained anatomical plausibility and automatic COCO-format ground-truth export;
(2)~a custom \texttt{PoseInjectorNode} feeds 3D-derived skeletons into a dual ControlNet (pose + depth) conditioned on FLUX-1~Dev, synthesizing 12{,}000 photorealistic images across 10 posture categories with low annotation drift;
(3)~ViTPose-based confidence filtering and targeted augmentation remove generation failures and improve robustness;
(4)~RTMPose-M (13.6M params) is fine-tuned on the synthetic data and paired with geometric feature engineering and a lightweight MLP for posture classification, then quantized to INT8 for real-time edge deployment.
On a real-child test set ($n{\approx}300$), the FP16 model achieves \textbf{71.2~AP}---a \textbf{+12.5~AP} improvement over the COCO-pretrained adult-data baseline at identical model capacity.
After INT8 quantization the model retains \textbf{70.4~AP} while running at \textbf{22~FPS} on a 0.8-TOPS Rockchip RK3568 NPU.
In a single-subject controlled comparison with a commercial posture corrector, our system achieves substantially higher recognition rates across most tested categories and responds \textbf{${\sim}$1.8$\times$~faster} on average.
These results demonstrate that carefully designed AIGC pipelines can substantially reduce dependence on real child imagery while achieving deployment-ready accuracy, with potential applications to other privacy-sensitive domains.

\vspace{0.5em}
\end{abstract}

\section{Introduction}
\label{sec:intro}

Poor sitting posture during prolonged study sessions is a growing health concern for school-age children, with epidemiological studies linking habitual postural deviations to musculoskeletal pain and spinal malalignment~\cite{calvo2013prevalence}, while close reading distance and prolonged near work are associated with myopia progression~\cite{ip2008nearwork}.
As children spend increasingly long hours at desks---both in school and at home---the demand for automated posture monitoring systems has given rise to a new category of consumer devices: AI-powered study companions that observe the child via a desk-mounted camera and issue real-time corrective alerts when problematic postures are detected.
The core technical challenge in building such devices is \emph{accurate child posture estimation under resource-constrained edge deployment}.

However, unlike adult pose estimation, which benefits from large-scale annotated benchmarks such as COCO~\cite{lin2014coco} and MPII~\cite{andriluka2014mpii}, the child posture domain faces a \textbf{fundamental data scarcity problem}.
Collecting and annotating large datasets of real children raises severe privacy and ethical concerns: minors cannot provide informed consent independently, institutional review requirements are stringent, and data leakage risks are amplified by the sensitive nature of home-environment imagery.
As a result, no public large-scale keypoint dataset of children in study scenarios currently exists, and commercial products typically rely on either (i)~adult-trained models that transfer poorly to children's distinct body proportions (as our experiments confirm in Section~\ref{sec:exp:main}), or (ii)~small proprietary datasets that lack the diversity to generalize across postures, body types, and environments.

A natural alternative is synthetic data generation.
Prior work has demonstrated that CG-rendered human models (e.g., SURREAL~\cite{varol2017surreal}) can provide unlimited annotated training data with perfect ground truth.
However, images produced by standard 3D rendering pipelines suffer from a well-documented \emph{sim-to-real gap}---the visual distribution mismatch between synthetic and real images---which severely degrades model performance when deployed on real photographs~\cite{tobin2017domain, nikolenko2021synthetic}.
Naive domain randomization partially mitigates this gap but cannot fully replicate the rich texture, lighting, and material properties of real-world scenes.
As a result, CG-only synthetic training remains substantially inferior to training on real data when real data is available.

In this paper, we propose an alternative strategy: rather than improving the CG renderer, we use the 3D model \emph{only} for its geometric ground truth and delegate photorealistic appearance synthesis to a state-of-the-art AI image generator.
Specifically, we present \textbf{\ours}, a four-stage AIGC-based synthetic data pipeline (Section~\ref{sec:method}) whose key insight is \emph{``3D geometry for truth, AIGC for realism''}---decoupling perfect annotations from a parametric body model and photorealistic appearance from a generative model~\cite{blackforestlabs2024flux}, without requiring any real child photographs.

\paragraph{Contributions.}
\begin{itemize}
  \item A complete AIGC synthetic data pipeline that generates photorealistic child posture images with ground-truth-projected keypoint annotations, requiring \emph{zero} real child photographs.
  \item A ground-truth pose injection mechanism (\texttt{PoseInjectorNode}) that achieves low annotation drift between the ControlNet conditioning signal and stored labels ($<\tau_{\text{drift}}{=}0.15$ normalized by bounding-box diagonal), verified by automated spatial-fidelity filtering (Section~\ref{sec:method:filter}).
  \item A complete training-to-deployment pipeline: fine-tuned RTMPose-M with geometric feature engineering and MLP posture classification, quantized to INT8 and running at 22~FPS on a 0.8-TOPS edge NPU.
  \item A pilot empirical comparison between an AIGC-synthetic-data-trained child posture system and a commercial product, providing preliminary evidence of competitive accuracy and lower latency.
\end{itemize}

\section{Related Work}
\label{sec:related}

\subsection{Human Pose Estimation}
\label{sec:related:pose}

Modern pose estimation architectures range from multi-resolution CNNs such as HRNet~\cite{sun2019hrnet} to Vision Transformers like ViTPose~\cite{xu2022vitpose}, all evaluated on adult benchmarks (COCO~\cite{lin2014coco}, MPII~\cite{andriluka2014mpii}).
On the efficiency frontier, RTMPose~\cite{jiang2023rtmpose} achieves real-time performance via a CSPNeXt backbone with SimCC-based coordinate classification, while OpenPose~\cite{cao2021openpose} established the skeleton visualization format now standard in pose-conditioned generation.
However, virtually all existing methods target \emph{adult} subjects, and the few studies addressing non-adult populations focus on infants~\cite{gama2024infant}, whose body proportions differ even more drastically from adults.
No public large-scale keypoint benchmark targets \emph{school-age children} in study environments.
The distinct anthropometric characteristics of this age group---a proportionally larger head-to-body ratio and shorter limbs compared to adults---cause adult-trained models to exhibit systematic errors, particularly on shoulder, elbow, and wrist keypoints.
Our experiments quantify this domain gap at $-$12.5~AP when a COCO-pretrained model is evaluated on child subjects (Section~\ref{sec:exp:main}).

\subsection{Synthetic Data for Human Pose}
\label{sec:related:synth}

SURREAL~\cite{varol2017surreal} demonstrated that SMPL~\cite{loper2015smpl} bodies rendered against random backgrounds can train pose estimators, and domain randomization~\cite{tobin2017domain} further improves sim-to-real transfer.
Wood \etal\cite{wood2021fakeit} showed synthetic face datasets can rival real data when sufficiently sophisticated.
Nevertheless, the sim-to-real gap persists~\cite{nikolenko2021synthetic}, and prior mitigations (style transfer, GAN refinement, mixed training) either require real data or introduce domain artifacts.
Our work takes a complementary approach: we use the CG pipeline \emph{solely} for geometric ground truth and delegate photorealistic appearance to a generative model (Section~\ref{sec:method:controlnet}).

\subsection{AIGC and Controllable Image Generation}
\label{sec:related:aigc}

Latent Diffusion Models~\cite{rombach2022stablediffusion} enabled high-resolution image synthesis, and ControlNet~\cite{zhang2023controlnet} introduced spatial conditioning (edge maps, pose skeletons, segmentation masks) that guides generation while preserving quality.
FLUX-1~\cite{blackforestlabs2024flux} further improves prompt adherence via a rectified flow transformer architecture.
Existing pose-conditioned data augmentation approaches typically \emph{re-estimate} poses from generated images (introducing annotation noise) or use single-condition ControlNet.
Our pipeline addresses both limitations with ground-truth pose injection and dual ControlNet conditioning, and is---to our knowledge---among the first to integrate multi-condition synthesis with a complete training-to-edge-deployment pipeline for child posture estimation.

Recent work has further explored diffusion models for generating annotated training data.
DatasetDM~\cite{wu2023datasetdm} extracts perception annotations directly from diffusion latent spaces, and several concurrent efforts combine 3D geometric control with diffusion synthesis to produce data with photorealistic appearance and geometric ground truth.
Our pipeline shares the high-level motivation of leveraging generative models for annotation-preserving synthesis, but targets the specific challenges of child posture estimation---requiring child-specific body models, desk-study pose templates, and edge-deployment constraints---and ensures label--conditioning consistency through deterministic ground-truth pose injection rather than latent-space annotation extraction or post-hoc re-estimation.

\section{Method}
\label{sec:method}

\begin{figure*}[t]
  \centering
  \includegraphics[width=\textwidth]{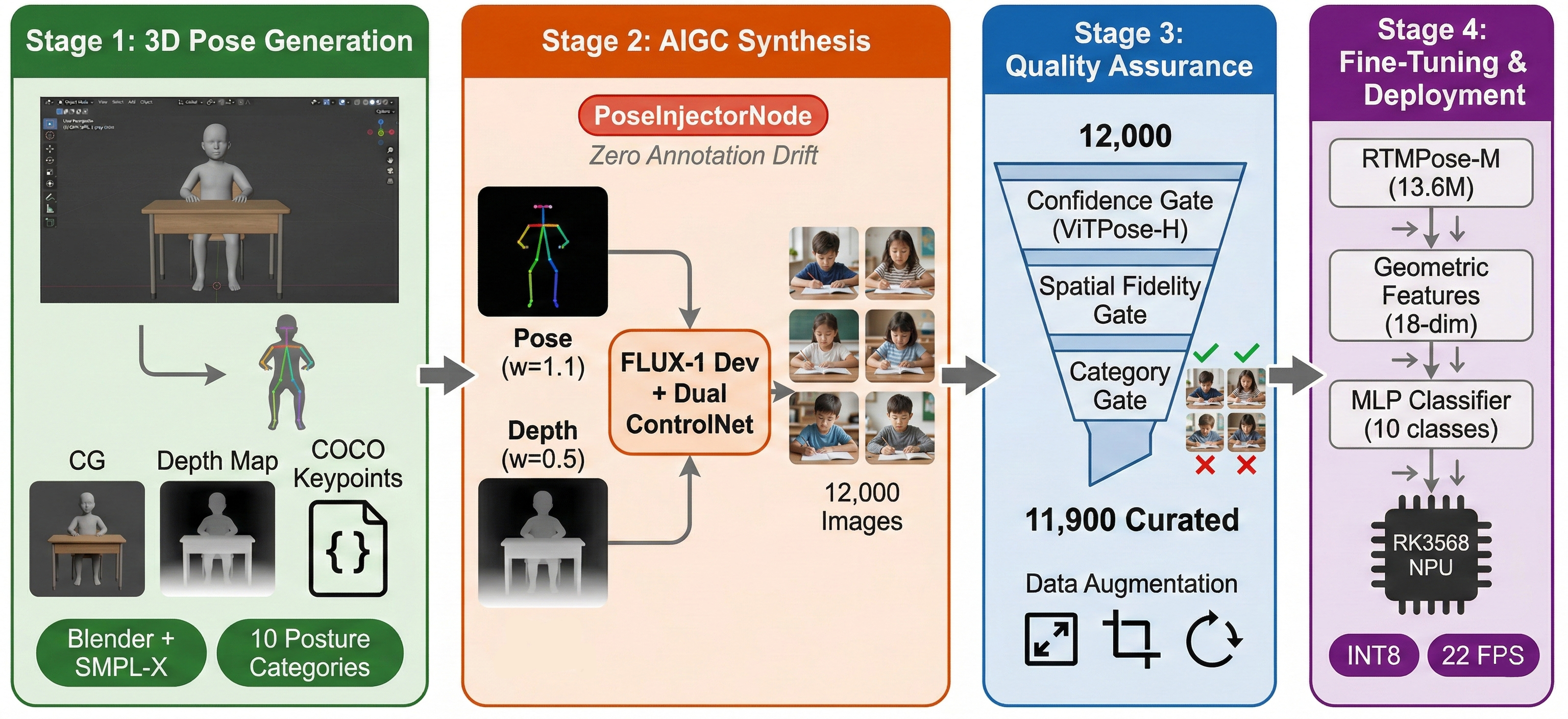}
  \caption{%
    Overview of the \ours pipeline.
    \textbf{Stage~1}: A child-specific SMPL-X model in Blender generates diverse desk-study poses with IK constraints, exporting rendered images, depth maps, and COCO-format keypoint annotations.
    \textbf{Stage~2}: A custom \texttt{PoseInjectorNode} feeds ground-truth skeletons into a dual ControlNet (pose + depth) conditioned on FLUX-1~Dev, producing 12{,}000 photorealistic child images with low annotation drift.
    \textbf{Stage~3}: Three automated quality gates---confidence filtering (ViTPose-H), spatial fidelity checking, and category consistency verification---remove generation failures and yield 11{,}900 curated images from an initial pool of 12{,}000; online augmentation further increases robustness.
    \textbf{Stage~4}: RTMPose-M is fine-tuned on the synthetic data; geometric feature engineering and a lightweight MLP classify postures; the model is quantized to INT8 for real-time inference on an edge NPU.%
  }
  \label{fig:pipeline}
\end{figure*}

Figure~\ref{fig:pipeline} provides an overview of the four-stage \ours pipeline. We describe each stage below.

\subsection{Stage~1: Programmable 3D Child Pose Generation}
\label{sec:method:3d}

\paragraph{Child Body Model.}
We adopt a modified SMPL-X body model~\cite{pavlakos2019smplx,loper2015smpl}, integrated into Blender via a custom plugin (\texttt{pose\_generator\_addon.py}) that supports child-specific body morphology adjustment and Rigify-based inverse kinematics.
The body mesh $M(\boldsymbol{\beta}, \boldsymbol{\theta})$ is parameterized by shape coefficients $\boldsymbol{\beta}\in\mathbb{R}^{10}$ and pose coefficients $\boldsymbol{\theta}\in\mathbb{R}^{J\times 3}$ for $J$ body joints, each with three rotation axes (we use the body-only subset of the SMPL-X kinematic tree).
Child-appropriate body proportions---shorter limbs, a larger head-to-body ratio, and narrower shoulders---are enforced through pre-calibrated $\boldsymbol{\beta}$ ranges tuned with the Blender plugin's anthropometric controls to match the 6--12 age bracket.

\paragraph{Inverse Kinematics Constraints.}
Unconstrained randomization of $\boldsymbol{\theta}$ inevitably produces anatomically implausible configurations (\eg, hyper-extended elbows or self-penetrating limbs).
We impose per-joint angular limits via the Rigify IK constraint system integrated into our Blender plugin.
For each joint $j$ along each rotation axis $a\in\{x,y,z\}$, the angle is bounded by
\begin{equation}
  \theta_{j,a}^{\min} \;\leq\; \theta_{j,a} \;\leq\; \theta_{j,a}^{\max},
  \label{eq:ik}
\end{equation}
where the bounds $(\theta_{j,a}^{\min},\, \theta_{j,a}^{\max})$ are empirically calibrated through iterative visual inspection and biomechanical plausibility testing.
Rather than adopting fixed textbook values, we found that task-specific tuning---focused on the range of upper-body motions observable in desk-study scenarios---yields more natural and diverse pose distributions.
Approximate angular bounds for key upper-body joints are: neck flexion/extension $\in[-40^{\circ},\, +15^{\circ}]$, shoulder abduction $\in[0^{\circ},\, +90^{\circ}]$, elbow flexion $\in[0^{\circ},\, +145^{\circ}]$; per-joint perturbation magnitudes $\epsilon_{j,a}$ range from $3^{\circ}$ to $12^{\circ}$.
Full per-joint specifications are provided in the released plugin configuration.

\paragraph{Action Templates and Pose Diversification.}
We define $C{=}10$ base posture templates corresponding to common desk-study behaviors, as summarized in Table~\ref{tab:poses}.
These encompass one correct sitting posture and nine deviation categories that collectively cover the postural problems most frequently observed in children's study environments.

\begin{table}[t]
  \centering
  \caption{The $C{=}10$ posture categories generated in Stage~1.
    Each template specifies a canonical joint configuration; stochastic perturbation (Eq.~\ref{eq:perturb}) generates diverse variants within each category.}
  \label{tab:poses}
  \small
  \begin{tabular}{@{}cl>{\raggedright\arraybackslash}p{4.8cm}@{}}
    \toprule
    \textbf{ID} & \textbf{Category} & \textbf{Description} \\
    \midrule
    0  & \texttt{correct\_posture} & Upright spine, shoulders level, eyes forward \\
    1  & \texttt{bowed\_head}      & Moderate forward head tilt (${\sim}15$--$30^{\circ}$) \\
    2  & \texttt{very\_bowed}      & Severe forward flexion ($>30^{\circ}$) \\
    3  & \texttt{lean\_desk}       & Upper body collapsed onto desk surface \\
    4  & \texttt{lean\_left}       & Lateral trunk lean to the left \\
    5  & \texttt{lean\_right}      & Lateral trunk lean to the right \\
    6  & \texttt{left\_headed}     & Head rotated toward left shoulder \\
    7  & \texttt{right\_headed}    & Head rotated toward right shoulder \\
    8  & \texttt{turn\_left}       & Upper body axial rotation leftward \\
    9  & \texttt{turn\_right}      & Upper body axial rotation rightward \\
    \bottomrule
  \end{tabular}
\end{table}

Each template $c$ specifies a canonical joint configuration $\boldsymbol{\theta}_{\text{base}}^{(c)}$, which is manually posed by the operator in Blender to ensure anatomical realism.
To generate intra-class diversity, we apply stochastic perturbation:
\begin{equation}
  \boldsymbol{\theta}^{(c,k)} = \boldsymbol{\theta}_{\text{base}}^{(c)} + \boldsymbol{\delta}^{(k)},
  \quad
  \delta_{j,a}^{(k)} \sim \mathcal{U}\!\left(-\epsilon_{j,a},\; \epsilon_{j,a}\right),
  \label{eq:perturb}
\end{equation}
subject to the IK constraints in Eq.~\eqref{eq:ik}, where $\epsilon_{j,a}$ controls the per-joint perturbation magnitude.
Any sample that violates a bound is clamped to the nearest valid angle.
This produces $N_{\text{var}}$ pose variants per template ($k = 1,\ldots, N_{\text{var}}$); crucially, all variants inherit the posture category label of their parent template, enabling the generated data to serve both keypoint regression and posture classification tasks.

\paragraph{Rendering and Annotation Export.}
For each pose $\boldsymbol{\theta}^{(c,k)}$, Blender renders a color image, a depth map from the Z-buffer, and a COCO-format JSON annotation file with 2D keypoint coordinates $\{(x_j, y_j, v_j)\}_{j=1}^{J}$ projected from the 3D joints, where $v_j$ follows the COCO convention ($0$: not labeled, $1$: labeled but occluded, $2$: labeled and visible).
In our pipeline, all rendered keypoints are set to $v_j{=}2$ (fully visible) since the canonical desk-study viewpoint does not produce severe self-occlusion for the selected upper-body joints.
All annotations are \emph{ground-truth by construction}, computed from the exact skeleton via known camera intrinsics.

\subsection{Stage~2: Multi-Condition Controllable Image Synthesis}
\label{sec:method:controlnet}

While Stage~1 produces geometrically accurate poses with perfect annotations, the rendered images exhibit a pronounced synthetic appearance that limits direct sim-to-real transfer.
Stage~2 bridges this domain gap by leveraging a state-of-the-art generative model to synthesize photorealistic images that faithfully preserve the 3D-derived pose structure.

\paragraph{Generative Backbone.}
We employ FLUX-1~Dev~\cite{blackforestlabs2024flux} in FP8 precision as the generation backbone, chosen for its superior prompt adherence and detail fidelity over prior latent diffusion models~\cite{rombach2022stablediffusion}.

\paragraph{Ground-Truth Pose Injection.}
A central design principle of our pipeline is the \emph{removal of external pose estimation preprocessors} during data generation.
We develop a custom ComfyUI node, \texttt{PoseInjectorNode}, that directly converts the COCO-format JSON annotations from Stage~1 into standardized OpenPose skeleton visualizations using deterministic rendering rules.
Formally, given keypoint coordinates $\{(x_j, y_j)\}_{j=1}^{J}$, the node renders:
\begin{equation}
  \mathbf{I}_{\text{pose}} = \mathcal{R}_{\text{OpenPose}}\!\left(\{(x_j, y_j)\}_{j=1}^{J}\right),
  \label{eq:render_pose}
\end{equation}
where $\mathcal{R}_{\text{OpenPose}}$ draws skeleton limbs and joint circles following the standard OpenPose color-coding convention.
This ensures \emph{low annotation drift} between the ControlNet conditioning signal and the stored ground-truth labels---a critical advantage over workflows that re-estimate poses from generated images using a separate detector, which inevitably introduces reprojection error.
Note that this guarantee concerns conditioning--label consistency; the fidelity of generated images to the conditioning signal is separately verified by the multi-criteria filter in Stage~3 (Section~\ref{sec:method:filter}).

\begin{figure}[t]
  \centering
  \includegraphics[width=\columnwidth]{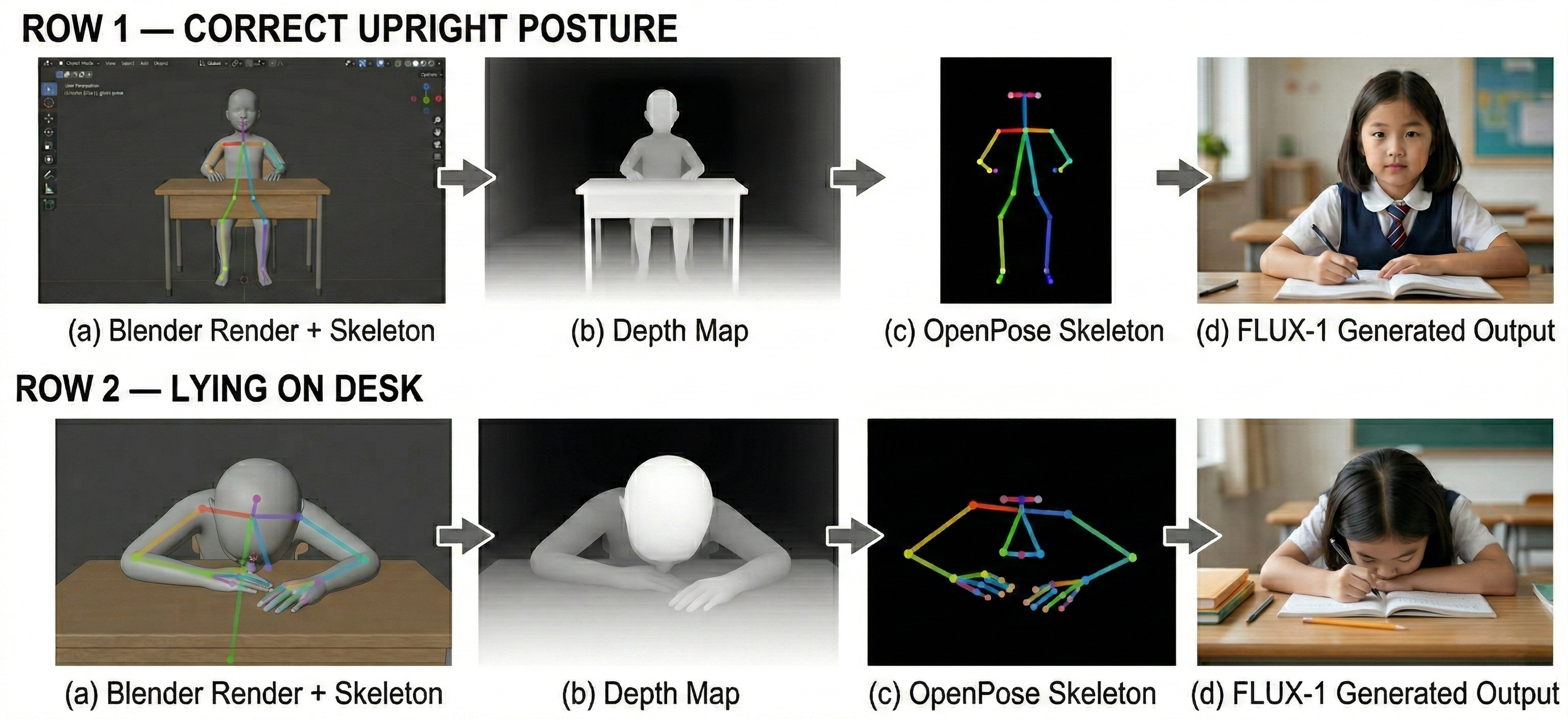}
  \caption{%
    Illustration of the Stage~1$\to$Stage~2 pipeline for two representative samples viewed from a frontal desk-mounted camera perspective.
    \emph{Row~1}: correct upright posture; \emph{Row~2}: upper body collapsed onto desk (\texttt{lean\_desk}).
    Each row shows:
    (a)~Blender render with overlaid 3D skeleton,
    (b)~depth map rendered from the Blender Z-buffer,
    (c)~COCO-17 OpenPose skeleton deterministically rendered by \texttt{PoseInjectorNode} from the ground-truth JSON, and
    (d)~photorealistic image generated by FLUX-1 conditioned on both control signals (pose + depth).
    The stark visual contrast between (a) and (d) demonstrates the pipeline's ability to bridge the sim-to-real gap while faithfully preserving the input pose.%
  }
  \label{fig:synthesis}
\end{figure}

\paragraph{Dual ControlNet Conditioning.}
To ensure both structural fidelity and visual realism, we condition the generative process through two complementary spatial control signals, each processed by a dedicated ControlNet adapter.
FLUX-1 employs a rectified-flow formulation in which a learned velocity field $\mathbf{v}_\phi$ transports noise $\mathbf{z}_1 \sim \mathcal{N}(\mathbf{0}, \mathbf{I})$ toward data $\mathbf{z}_0$ along the interpolation path $\mathbf{z}_t = (1{-}t)\,\mathbf{z}_0 + t\,\mathbf{z}_1$.
The multi-condition generation augments this velocity field with weighted residuals from two ControlNet branches:
\begin{equation}
  \mathbf{v}_\phi(\mathbf{z}_t,\, t,\, \mathbf{c}_{\text{text}})
  \;\mathrel{+}=\;
  \sum_{i=1}^{2} w_i \cdot \mathcal{F}_{\text{CN}}^{(i)}\!\left(\mathbf{z}_t,\, \mathbf{h}_i,\, t\right),
  \label{eq:controlnet}
\end{equation}
where $\mathbf{z}_t$ is the interpolated latent at time $t \in [0, 1]$, $\mathcal{F}_{\text{CN}}^{(i)}$ is the $i$-th ControlNet branch, and the control inputs $\{\mathbf{h}_i\}$ with corresponding weights $\{w_i\}$ are:
\begin{enumerate}
  \item \textbf{Pose control} ($w_1{=}1.1$): The OpenPose skeleton $\mathbf{I}_{\text{pose}}$ (Eq.~\ref{eq:render_pose}), enforcing joint-level body structure with above-unity weight to strongly anchor the generated pose.
  \item \textbf{Depth control} ($w_2{=}0.5$): The depth map $\mathbf{I}_{\text{depth}}$ rendered from the 3D scene in Stage~1, providing geometric structure cues---body volume, spatial layering, and figure--background separation---that guide the generator toward plausible 3D-consistent appearances.
\end{enumerate}
The asymmetric weighting ($w_1 > w_2$) prioritizes pose fidelity---the most critical property for downstream keypoint supervision---while the depth channel anchors spatial composition without over-constraining surface appearance.
The pose adapter is sourced from the XLabs-AI FLUX ControlNet collection, and the depth adapter from the FLUX.1-dev-Depth control model released by Black Forest Labs~\cite{blackforestlabs2024flux}.
We use the term ``ControlNet'' loosely to encompass both spatial-conditioning mechanisms, noting that FLUX-1's transformer-based architecture may implement depth conditioning differently from the encoder-copy design of the original ControlNet~\cite{zhang2023controlnet}.

\paragraph{Combinatorial Prompt Engineering.}
To maximize visual diversity across the generated dataset, we construct text prompts through structured random composition over multiple independent attribute dimensions:
\begin{equation}
  p = f_{\text{concat}}\!\left(
    \underbrace{a}_{\text{age}},\;
    \underbrace{g}_{\text{gender}},\;
    \underbrace{b}_{\text{body}},\;
    \underbrace{h}_{\text{hair}},\;
    \underbrace{s}_{\text{clothing}},\;
    \underbrace{e}_{\text{background}}
  \right),
  \label{eq:prompt}
\end{equation}
where each attribute is independently sampled from a curated vocabulary (\eg, age, gender, clothing style, background).
A fixed negative prompt suppresses common artifacts.
This combinatorial strategy yields $\prod_{d} |\mathcal{V}_d|$ unique appearance configurations.
Across all $C{=}10$ categories, we generate 12{,}000 images (${\sim}$1{,}200 per class).

\subsection{Stage~3: Automated Quality Filtering and Augmentation}
\label{sec:method:filter}

Despite careful dual conditioning, a fraction of generated images exhibit structural failures---misaligned limbs, missing body parts, or poses that do not match the intended category.
We apply a two-tier filtering protocol followed by online data augmentation.

\paragraph{Automated Confidence-Based Filtering.}
We run a pretrained ViTPose-H model~\cite{xu2022vitpose} on each generated image $\mathbf{x}$ and obtain re-estimated keypoints $\{\hat{\mathbf{y}}_j\}_{j=1}^{J}$ with per-keypoint confidence scores $\{s_j\}$.
An image is accepted only if it passes all three gates of a multi-criteria filter:
\begin{equation}
  \mathbf{x} \in \mathcal{D}_{\text{accept}}
  \;\iff\;
  \underbrace{\min_{j} s_j > \tau_{\text{conf}}}_{\text{confidence gate}}
  \;\wedge\;
  \underbrace{\max_{j} \frac{\|\hat{\mathbf{y}}_j - \mathbf{y}_j^{\text{gt}}\|_2}{\sqrt{w^2+h^2}} < \tau_{\text{drift}}}_{\text{spatial fidelity gate}}
  \;\wedge\;
  \underbrace{c_{\text{pred}} = c_{\text{target}}}_{\text{category gate}},
  \label{eq:filter}
\end{equation}
where $\mathbf{y}_j^{\text{gt}}$ are the ground-truth keypoints from Stage~1, $\sqrt{w^2{+}h^2}$ is the bounding-box diagonal used for scale normalization, $c_{\text{pred}}$ is the posture class inferred from the re-estimated skeleton, and $c_{\text{target}}$ is the intended label.
We set $\tau_{\text{conf}}{=}0.5$ and $\tau_{\text{drift}}{=}0.15$ based on a small validation subset of 200~images.
The bounding box $(w, h)$ used for scale normalization is the tight enclosing rectangle of all ground-truth keypoints $\mathbf{y}_j^{\text{gt}}$ from Stage~1.
The confidence gate removes catastrophic generation failures (phantom limbs, missing torsos); the spatial fidelity gate catches images where the synthesized body deviates from the conditioning skeleton despite appearing superficially plausible; and the category gate discards borderline poses whose re-estimated class disagrees with the intended label.

\paragraph{Manual Review.}
A brief manual pass reviews borderline cases near the automated thresholds and discards residual artifacts (ambiguous category boundaries, rare generation quirks).
Together with the automated gates, the full quality-assurance pipeline removes approximately 100 images in total from the initial pool, yielding a final curated set of \textbf{11{,}900 annotated images}.

\paragraph{Empty-Scene Detection.}
For deployment, seat-vacancy detection is handled by a separate lightweight person-detection model that operates independently of the posture estimation and classification pipeline described above.
This module is not evaluated in the controlled experiments of Section~\ref{sec:exp}.

\paragraph{Data Augmentation.}
We apply three online augmentations during training to simulate edge-device conditions:
\begin{itemize}
  \item \textbf{Resolution jittering}: downscale by $r_{\text{scale}} \sim \mathcal{U}(0.5, 1.0)$ then upscale, simulating low-quality camera feeds.
  \item \textbf{Random cropping}: scale $s_{\text{crop}} \sim \mathcal{U}(0.8, 1.0)$ with random offset, emulating imperfect framing.
  \item \textbf{Micro-rotation}: $\phi \sim \mathcal{U}(-5^{\circ}, 5^{\circ})$, accounting for tilted camera mounting.
\end{itemize}

\subsection{Stage~4: Posture Classification and Edge Deployment}
\label{sec:method:deploy}

The target deployment platform---an edge-class neural processing unit (NPU)---imposes strict constraints on model size, memory footprint, and inference latency.
We adopt a two-stage inference architecture: a lightweight pose estimator extracts keypoints from each frame, followed by geometric feature engineering and an MLP classifier for posture categorization.

\paragraph{Pose Estimator Fine-Tuning.}
We fine-tune RTMPose-M~\cite{jiang2023rtmpose} (13.6M params, $256{\times}192$ input) on our synthetic dataset using MMPose~\cite{mmpose2020}, initializing from COCO-pretrained weights to leverage learned adult-pose representations before domain adaptation to child subjects.
RTMPose-M uses a CSPNeXt backbone with SimCC-based coordinate classification~\cite{jiang2023rtmpose}, which directly regresses discretized $x$- and $y$-coordinates rather than producing dense heatmaps, yielding a compact architecture well-suited for NPU deployment.

\paragraph{Geometric Feature Engineering.}
From the $K{=}17$ COCO-format keypoints predicted by RTMPose-M, we first select $K'{=}13$ upper-body joints relevant to seated posture (discarding ankle and knee keypoints rarely visible in desk-study framing).
From these, we construct a compact feature vector $\mathbf{f} \in \mathbb{R}^{18}$ by concatenating two groups:
\begin{equation}
  \mathbf{f} = \left[\, \underbrace{y_1',\; y_2',\; \ldots,\; y_{13}'}_{\text{13 normalized joint coordinates}},\;
  \underbrace{\alpha_{\text{spine}},\; \alpha_{\text{head}},\; r_{\text{shoulder}},\; h_{\text{eye}},\; d_{\text{lateral}}}_{\text{5 geometric quantities}} \,\right]^\top,
  \label{eq:features}
\end{equation}
where $y_j'$ denotes the z-score normalized vertical coordinate of the $j$-th selected upper-body joint ($j = 1, \ldots, K'{=}13$), capturing the spatial layout of the upper body in a pose-estimation-agnostic form.
The five geometric quantities encode task-specific postural semantics:
$\alpha_{\text{spine}}$ is the spine inclination angle (shoulder-midpoint to hip-midpoint deviation from vertical),
$\alpha_{\text{head}}$ is the head tilt relative to the shoulder line,
$r_{\text{shoulder}}$ is the shoulder tilt ratio (left--right height difference normalized by shoulder width),
$h_{\text{eye}}$ is the eye-to-shoulder vertical distance capturing head-bowing severity, and
$d_{\text{lateral}}$ measures trunk lateral displacement.
All 18 features are z-score normalized using per-feature statistics computed from the training set.
This two-group design separates raw spatial layout (which the MLP can learn to exploit freely) from interpretable geometric priors aligned with the posture categories in Table~\ref{tab:poses}.

\paragraph{MLP Posture Classifier.}
A lightweight multi-layer perceptron maps the 18-dimensional feature vector to $C{=}10$ posture classes (Table~\ref{tab:poses}).
The network comprises two hidden layers (128 and 64~units), each followed by batch normalization, ReLU activation, and dropout ($p{=}0.3$):
\begin{equation}
  \hat{y} = \mathbf{W}_3\,\mathbf{h}_2 + \mathbf{b}_3,
  \quad
  \mathbf{h}_l = \operatorname{drop}_{p}\!\left(\sigma\!\left(\operatorname{BN}\!\left(\mathbf{W}_l\,\mathbf{h}_{l-1} + \mathbf{b}_l\right)\right)\right),
  \quad
  \mathbf{h}_0 = \mathbf{f},
  \label{eq:mlp}
\end{equation}
where $\operatorname{BN}$ denotes batch normalization, $\sigma$ is ReLU, and $\operatorname{drop}_p$ is dropout with rate $p{=}0.3$.
The MLP totals fewer than 12K trainable parameters---negligible overhead relative to the 13.6M-parameter pose estimator.

\paragraph{Quantization and Deployment.}
The fine-tuned RTMPose-M backbone and MLP classifier are quantized to INT8 for the target Rockchip RK3568 NPU (0.8~TOPS), enabling real-time on-device posture monitoring without cloud connectivity.

\section{Experiments}
\label{sec:exp}

We evaluate our pipeline along three complementary dimensions:
(i)~standard keypoint estimation metrics on a real-child test set (Section~\ref{sec:exp:main}),
(ii)~end-to-end posture recognition accuracy and response latency in a pilot comparison with a commercial product (Section~\ref{sec:exp:e2e}), and
(iii)~ablation studies isolating the contribution of each pipeline component (Section~\ref{sec:exp:ablation}).

\subsection{Experimental Setup}
\label{sec:exp:setup}

\paragraph{Synthetic Training Set.}
As described in Section~\ref{sec:method}, our pipeline produces 11{,}900 curated synthetic images spanning $C{=}10$ posture categories, each annotated with COCO-format keypoints.
The dataset is split into 10{,}700 training images and 1{,}200 validation images with stratified sampling to preserve class balance.

\paragraph{Real-Child Test Set.}
To evaluate generalization to real imagery, we collect a private test set of approximately 300 images captured in authentic home study environments.
The subjects are four children (ages 7--12) photographed during natural study sessions across four distinct home environments, each using a desk-mounted camera at $D{\approx}51$\,cm---the same deployment distance as the target edge device.
All subjects fall within the 6--12 target age range of our synthetic pipeline.
Informed consent was obtained from all participants' guardians.
All images are manually annotated with COCO keypoints by two independent annotators; inter-annotator agreement (PCK@0.2) exceeds 95\%, and the final labels are obtained by averaging.
The test set covers all 10 posture categories with emphasis on the more challenging deviation poses (\eg, lateral leaning, head bowing).
\emph{No images from this set are used during training.}

\paragraph{Baseline.}
As a reference point, we evaluate the COCO-pretrained RTMPose-M~\cite{jiang2023rtmpose} checkpoint---the same initialization used by our pipeline---directly on the child test set without any domain-specific fine-tuning.
This represents the best available adult-trained model at identical capacity and isolates the contribution of our synthetic child data.
We also conduct a pilot comparison with a commercial RGB-based posture corrector under identical conditions.

\paragraph{Evaluation Metrics.}
For keypoint-level evaluation, we report \textbf{Average Precision (AP)} following the COCO evaluation protocol and \textbf{PCK@0.2} (Percentage of Correct Keypoints within 20\% of the head segment length).
For end-to-end evaluation, we report \textbf{Recognition Rate (RR)}—the proportion of test instances where the system correctly triggers a posture alert within a 30-second window—and \textbf{Average Response Time ($T_{\text{avg}}$)}—the mean latency from posture onset to alert trigger.

\paragraph{Training Details.}
All models are trained on a single NVIDIA RTX~5090 GPU.
RTMPose-M~\cite{jiang2023rtmpose} is initialized from COCO-pretrained weights and fine-tuned for 270 epochs using AdamW~($\beta_1{=}0.9$, $\beta_2{=}0.999$) with an initial learning rate of $5{\times}10^{-4}$ and cosine annealing, batch size~128, input resolution $256{\times}192$.
Standard MMPose~\cite{mmpose2020} data loading and preprocessing pipelines are used throughout.
The MLP posture classifier is trained for 100 epochs with Adam~($\text{lr}{=}10^{-3}$) and cross-entropy loss on the geometric features extracted from the training set.
The converged models are converted to RKNN format and quantized to INT8 for deployment on the Rockchip RK3568 NPU (0.8~TOPS).
All reported FPS figures reflect \emph{end-to-end} throughput including image decoding, preprocessing (resize, normalization), NPU inference, and postprocessing (keypoint decoding, feature extraction, MLP classification), measured as the wall-clock average over 200 consecutive frames on the target hardware.

\subsection{Main Results: Keypoint Estimation}
\label{sec:exp:main}

Table~\ref{tab:main} presents the keypoint estimation performance of each training-data configuration on the real-child test set.

\begin{table}[t]
  \centering
  \caption{%
    Keypoint estimation accuracy on the real-child test set ($n{\approx}300$).
    All baseline and proposed models use the RTMPose-M architecture (13.6M params).
    ``$\dagger$'' denotes INT8 quantized model; all other rows use FP16 precision.
    All models are deployed on the same RK3568 NPU.
    Best results are in \textbf{bold}.%
  }
  \label{tab:main}
  \small
  \begin{tabular}{@{}llcccc@{}}
    \toprule
    \textbf{Training Data} & \textbf{Model} & \textbf{Params} & \textbf{AP} & \textbf{PCK@0.2} & \textbf{FPS} \\
    & & \textbf{(M)} & & & \textbf{(NPU)} \\
    \midrule
    COCO pre-trained       & RTMPose-M & 13.6 & 58.7 & 71.4 & 18 \\
    \midrule
    \ours                  & RTMPose-M & 13.6 & \textbf{71.2} & \textbf{84.5} & 18 \\
    \ours (INT8$\dagger$)  & RTMPose-M & 13.6 & 70.4 & 83.7 & \textbf{22} \\
    \bottomrule
  \end{tabular}
\end{table}

Several observations emerge from Table~\ref{tab:main}:

\noindent\textbf{(1)~Domain-specific synthetic data substantially outperforms adult real data.}
The COCO-pretrained model achieves only 58.7~AP on child subjects, reflecting the well-documented anthropometric domain gap between adult training data and children's distinct body proportions (larger head-to-body ratio, shorter limbs).
Fine-tuning on our AIGC synthetic child data closes this gap by \textbf{+12.5~AP}, suggesting that domain-appropriate synthetic data can outperform domain-mismatched real data even when no real child photographs are available for training.
These gains, while directionally consistent across all tested categories, should be interpreted as preliminary given the modest test-set scale (4~subjects; Section~\ref{sec:exp:setup}).

\noindent\textbf{(2)~INT8 quantization incurs minimal accuracy loss.}
The quantized model retains 98.9\% of the FP16 model's AP (70.4 vs.\ 71.2) while increasing NPU throughput from 18 to 22~FPS, confirming deployment readiness.

\paragraph{Statistical Caveat.}
Given the modest test-set size ($n{\approx}300$, 4~subjects), the reported AP and RR figures should be interpreted as point estimates without formal confidence intervals.
We caution against over-interpreting small absolute differences and emphasize the directional consistency of improvements across all metrics and categories.

\subsection{Pilot Comparison with a Commercial Product}
\label{sec:exp:e2e}

As a preliminary point of reference, we conduct a single-subject controlled comparison between our system and a commercial RGB-based children's posture corrector (a top-selling device on the Chinese consumer market, equipped with a single RGB camera and on-device processing; it supports head-turning, head-too-low, lateral-leaning, and lying-on-desk detection at factory-default sensitivity settings).
Both devices are placed at $D{\approx}51$\,cm on the same desk under controlled illumination (300--500~lux) with the same test subject.
\emph{This pilot comparison provides a directional indication of relative performance rather than a statistically definitive conclusion}; multi-subject validation is an important direction for future work (Section~\ref{sec:discussion:limitations}).

\paragraph{Recognition Rate.}
Table~\ref{tab:e2e_rr} presents the posture recognition accuracy across five evaluated behaviors.
Each behavior is tested at multiple angles or states with $N{=}3$ repetitions per condition.

\begin{table}[t]
  \centering
  \caption{%
    Pilot recognition rate (RR) comparison with a commercial RGB-based posture corrector (single subject, single environment).
    RR is the proportion of test instances where the device triggers a correct alert within 30\,s.
    ``---'' indicates the device does not support this detection category.%
  }
  \label{tab:e2e_rr}
  \small
  \begin{tabular}{@{}lccccc@{}}
    \toprule
    \textbf{Posture Behavior} & \textbf{Angle/Condition} & $n$ & \textbf{Ours} & \textbf{Commercial} & $\boldsymbol{\Delta}$ \\
    \midrule
    Head turning       & $15^{\circ}$--$50^{\circ}$ ($5^{\circ}$ steps) & 24 & 87.5\% & 83.3\% & \textbf{+4.2\%} \\
    Head too low       & $20^{\circ}$--$60^{\circ}$ ($10^{\circ}$ steps) & 15 & 86.7\% & 6.7\% & \textbf{+80.0\%} \\
    Lateral leaning    & $10^{\circ}$--$40^{\circ}$ ($5^{\circ}$ steps) & 21 & 66.7\% & 47.6\% & \textbf{+19.1\%} \\
    Body rotation      & $15^{\circ}$--$50^{\circ}$ ($5^{\circ}$ steps) & 24 & 91.7\% & --- & --- \\
    Lying on desk      & Fully collapsed & 3  & 100\%  & 100\% & 0\% \\
    \bottomrule
  \end{tabular}
\end{table}

The most notable gap appears on ``head too low,'' where the commercial product detects only 6.7\% of instances versus 86.7\% for our system---a difference that may partly reflect differing sensitivity thresholds rather than a purely algorithmic advantage.
Our system also uniquely supports \textbf{body rotation} detection (91.7\% RR), a category absent from the commercial product.

\paragraph{MLP Classification Accuracy.}
Evaluated separately on the real-child test set, the MLP posture classifier achieves 86.8\% overall accuracy and 0.84 macro-F1 across the deployed posture categories, with per-class precision ${\geq}0.73$ for all categories.
Classification errors concentrate on the boundary between \textit{moderate} and \textit{severe} head bowing, where angular differences are small.

\paragraph{Response Latency.}
Table~\ref{tab:e2e_latency} reports the average end-to-end response time—from posture onset to alert trigger—for all successfully detected instances.

\begin{table}[t]
  \centering
  \caption{%
    Average response time $T_{\text{avg}}$ (seconds) from posture onset to alert trigger.
    Per-category averages are computed over successfully detected instances; $n$ denotes the total number of test instances per category (same as Table~\ref{tab:e2e_rr}).
    Lower is better. ``---'' indicates the device does not support this category.
    The weighted average is computed over shared categories only.%
  }
  \label{tab:e2e_latency}
  \small
  \begin{tabular}{@{}lcccc@{}}
    \toprule
    \textbf{Posture Behavior} & $n$ & \textbf{Ours (s)} & \textbf{Commercial (s)} & \textbf{Speedup} \\
    \midrule
    Head turning       & 24 & \textbf{4.71} & 10.20 & $2.16\times$ \\
    Head too low       & 15 & \textbf{9.08} & 10.00 & $1.10\times$ \\
    Lateral leaning    & 21 & \textbf{6.50} & 13.70 & $2.11\times$ \\
    Body rotation      & 24 & \textbf{5.80} & ---   & --- \\
    Lying on desk      & 3  & \textbf{6.33} & 10.00 & $1.58\times$ \\
    \midrule
    \textbf{Weighted avg.} (shared) & 63 & \textbf{6.42} & 11.31 & $1.76\times$ \\
    \bottomrule
  \end{tabular}
\end{table}

Across the four shared categories, our system responds \textbf{$\boldsymbol{1.76{\times}}$ faster} on average, with the largest speedups on \textit{head turning} ($2.16{\times}$) and \textit{lateral leaning} ($2.11{\times}$)---subtle shifts where rapid detection is most valuable.

\subsection{Ablation Studies}
\label{sec:exp:ablation}

We conduct ablation experiments to quantify the contribution of each pipeline component.
All ablations use the RTMPose-M architecture and are evaluated on the real-child test set.

\begin{table}[t]
  \centering
  \caption{%
    Ablation study on key pipeline components.
    All variants use RTMPose-M evaluated on the real-child test set.
    $\Delta$\,AP denotes the change relative to the full pipeline.%
  }
  \label{tab:ablation}
  \small
  \begin{tabular}{@{}lccr@{}}
    \toprule
    \textbf{Configuration} & \textbf{AP} & \textbf{PCK@0.2} & $\boldsymbol{\Delta}$\textbf{AP} \\
    \midrule
    Full \ours pipeline                  & \textbf{71.2} & \textbf{84.5} & --- \\
    \midrule
    \textit{(a) Synthesis configuration} \\
    \quad w/o depth ControlNet ($w_2{=}0$)          & 68.1 & 81.5 & $-$3.1 \\
    \midrule
    \textit{(b) Quality control} \\
    \quad w/o ViTPose confidence filtering        & 67.8 & 81.2 & $-$3.4 \\
    \quad w/o data augmentation                   & 69.1 & 82.6 & $-$2.1 \\
    \midrule
    \textit{(c) Training strategy} \\
    \quad w/o COCO pre-trained initialization     & 65.3 & 78.1 & $-$5.9 \\
    \quad Frozen backbone (head-only fine-tuning) & 67.9 & 81.0 & $-$3.3 \\
    \bottomrule
  \end{tabular}
\end{table}

\paragraph{Analysis.}
\textbf{(a)}~We do not include a CG-only baseline (training directly on raw Blender renders) because preliminary experiments yielded AP substantially below the COCO-pretrained baseline, consistent with the severe sim-to-real gap documented in prior work~\cite{tobin2017domain, nikolenko2021synthetic}; a formal CG-only evaluation is deferred to future work.
Within our AIGC pipeline, disabling the depth ControlNet branch costs $-$3.1~AP, confirming that geometric structure cues provide meaningful complementary guidance beyond pose conditioning alone.
\textbf{(b)}~Disabling confidence filtering costs $-$3.4~AP despite removing only ${\sim}$0.8\% of images; this outsized impact is attributable to catastrophic generation failures (severely displaced bodies or missing limbs) that introduce maximally erroneous gradient signals during training. Augmentation contributes $-$2.1~AP.
\textbf{(c)}~Without COCO pre-trained initialization, accuracy drops by $-$5.9~AP, confirming that adult-pose priors provide a valuable warm start for child-domain adaptation.
Freezing the backbone and fine-tuning only the prediction head costs $-$3.3~AP, indicating that feature-level adaptation to child body proportions is important.

\subsection{Qualitative Observations}
\label{sec:exp:qualitative}

\begin{figure*}[t]
  \centering
  \setlength{\tabcolsep}{1.5pt}
  \begin{tabular}{@{}cccc@{}}
    {\small COCO Baseline} & {\small Ours (\ours)} &
    {\small COCO Baseline} & {\small Ours (\ours)} \\[2pt]
    \includegraphics[width=0.235\textwidth]{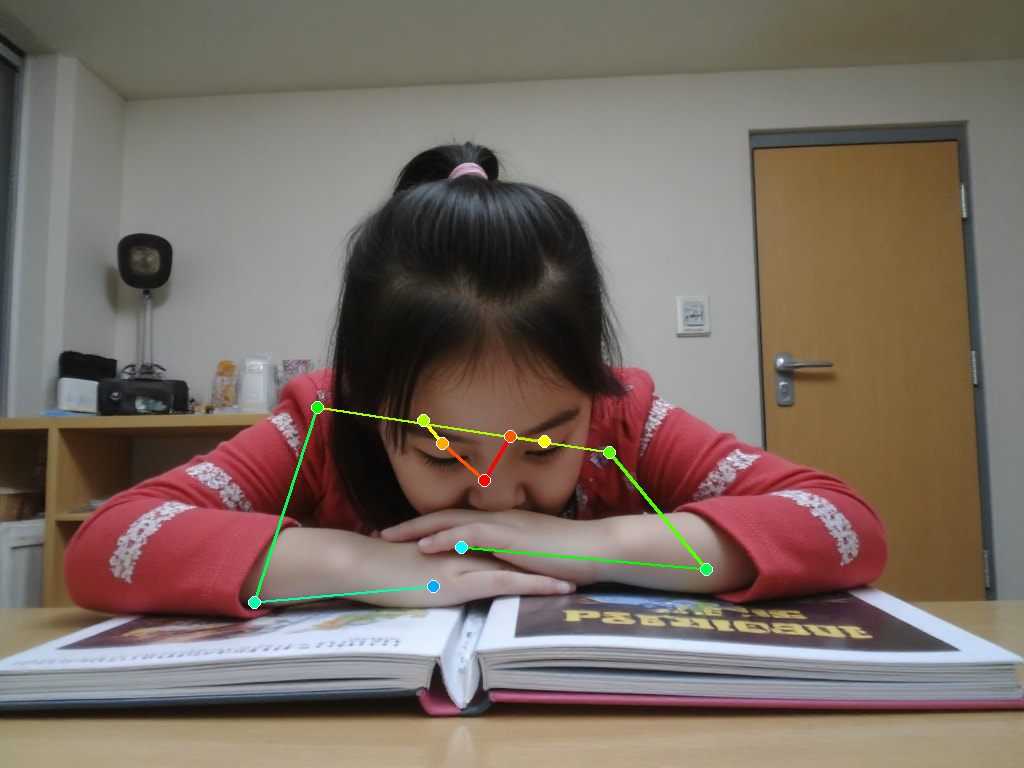} &
    \includegraphics[width=0.235\textwidth]{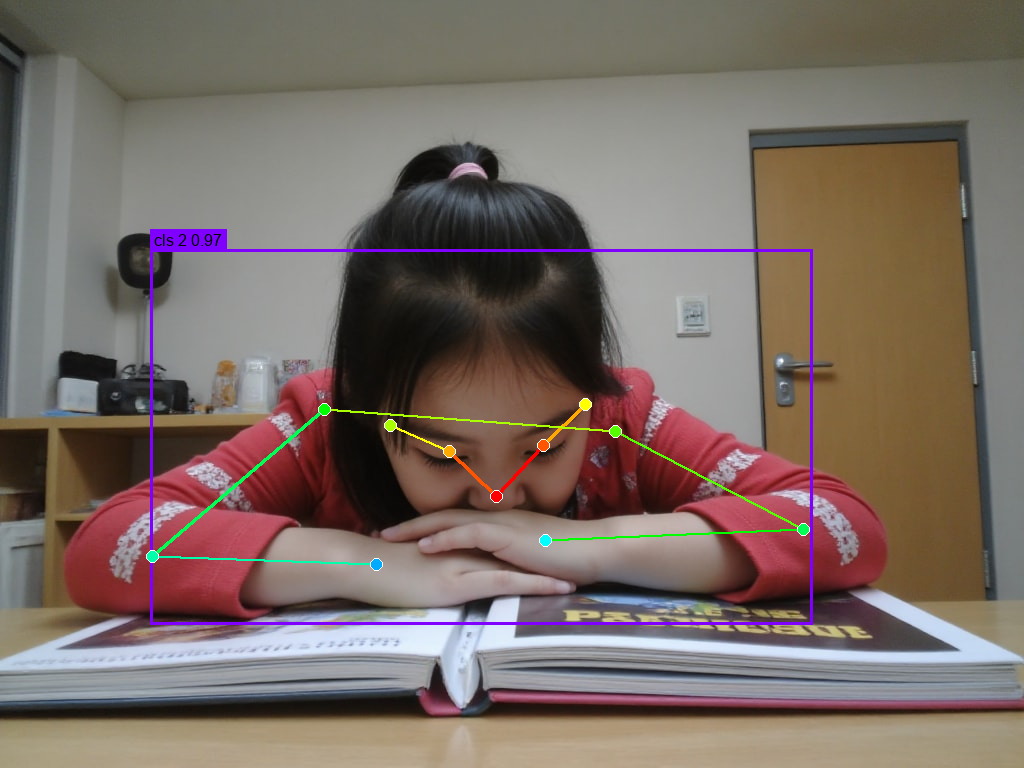} &
    \includegraphics[width=0.235\textwidth]{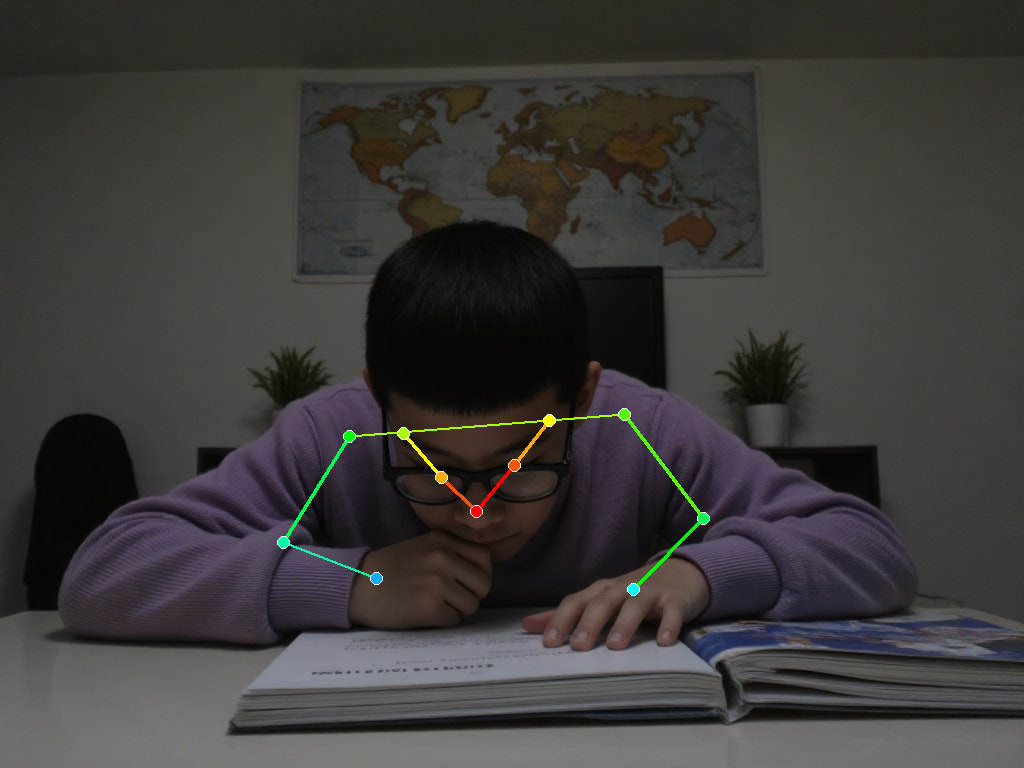} &
    \includegraphics[width=0.235\textwidth]{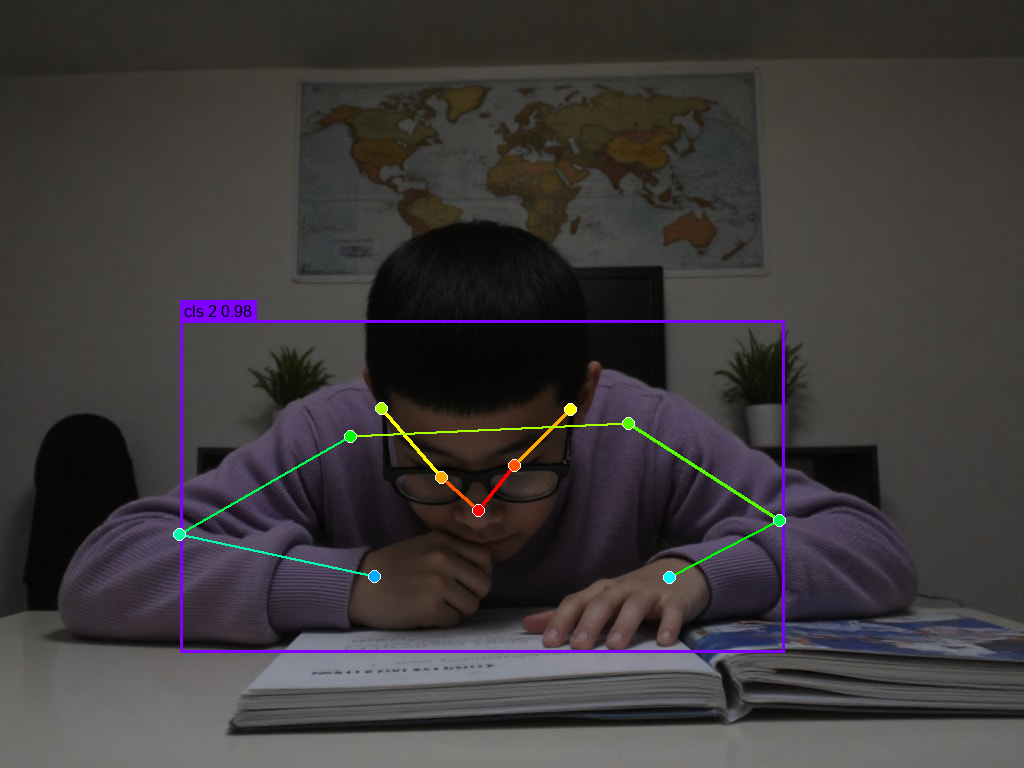} \\
    \multicolumn{2}{c}{\small (a)~Desk leaning---Subject~A} &
    \multicolumn{2}{c}{\small (b)~Desk leaning---Subject~B} \\[5pt]
    \includegraphics[width=0.235\textwidth]{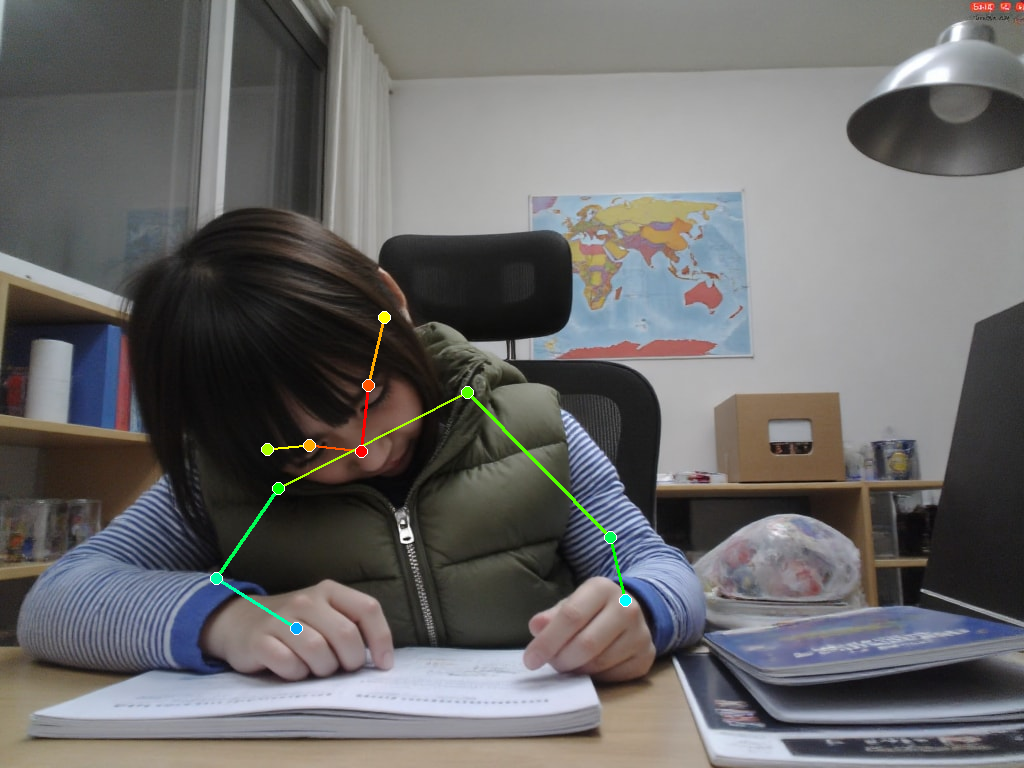} &
    \includegraphics[width=0.235\textwidth]{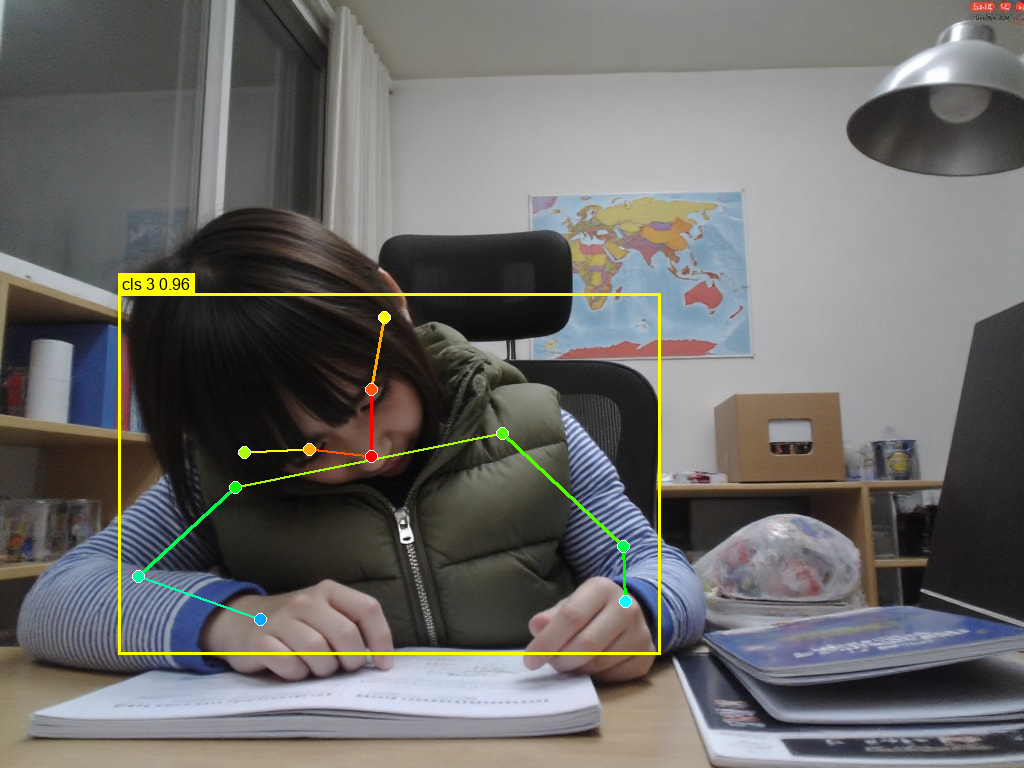} &
    \includegraphics[width=0.235\textwidth]{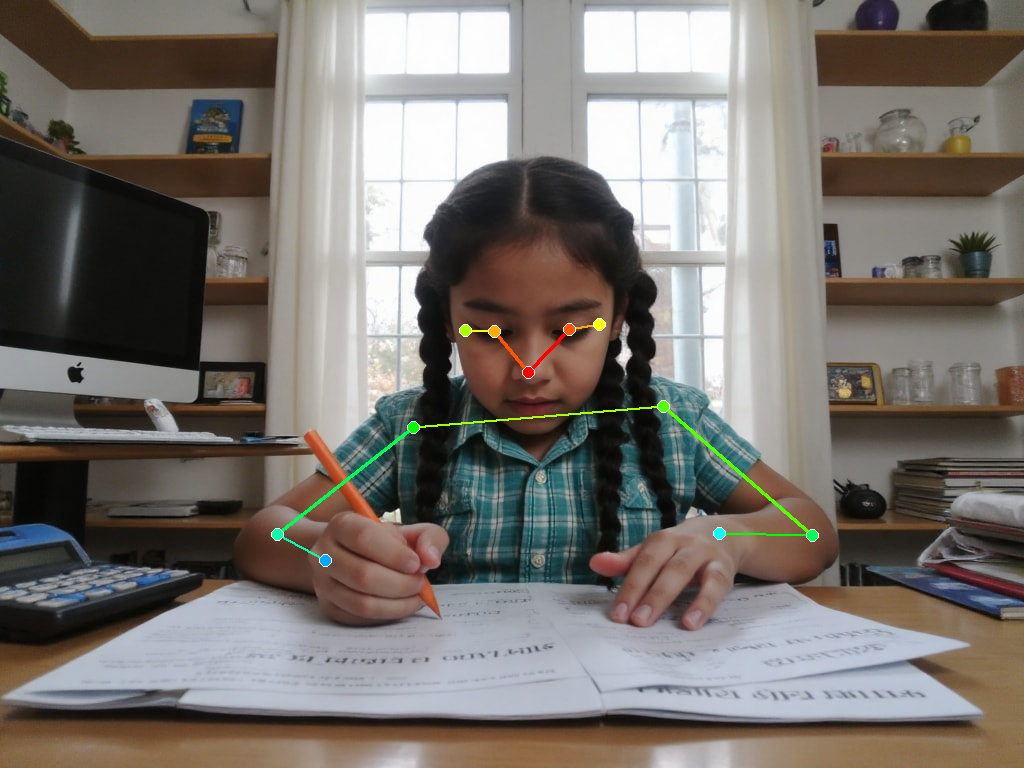} &
    \includegraphics[width=0.235\textwidth]{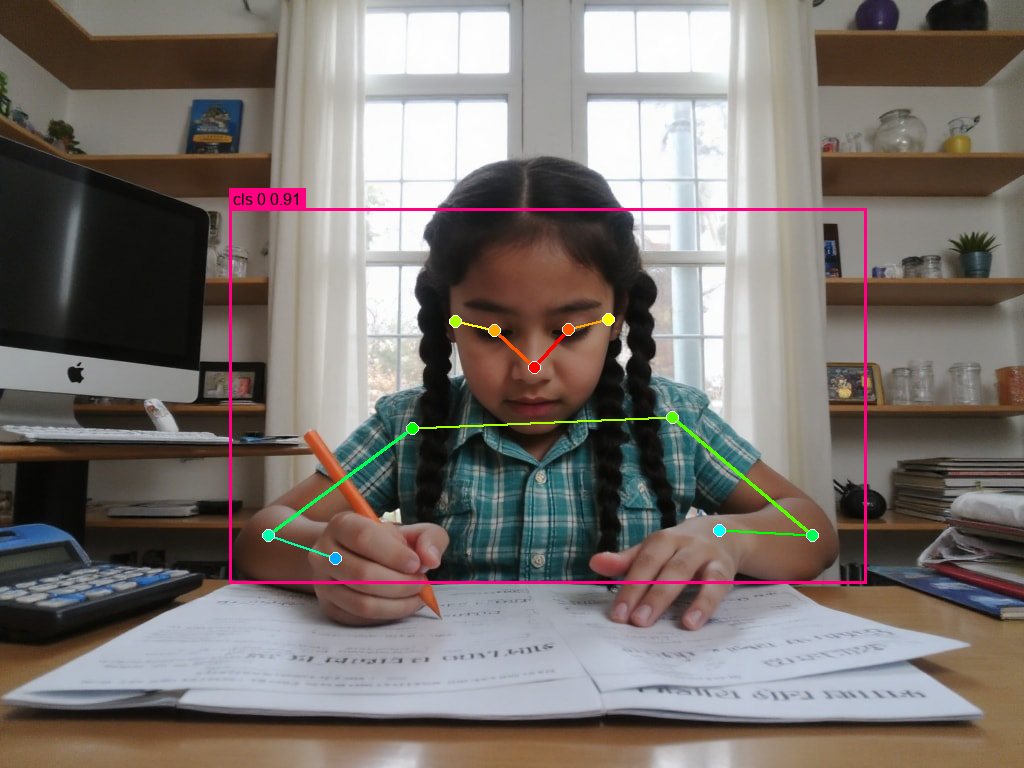} \\
    \multicolumn{2}{c}{\small (c)~Lean right} &
    \multicolumn{2}{c}{\small (d)~Correct upright posture} \\
  \end{tabular}
  \caption{%
    Qualitative comparison on challenging child postures from the real-child test set,
    selected for their difficulty under the COCO-pretrained baseline.
    Each pair shows the baseline output (\emph{left}: estimated keypoint skeleton only)
    and the full \ours pipeline output (\emph{right}: keypoints + detection bounding box
    + MLP classification label with confidence score).
    (a,\,b)~Two subjects leaning onto the desk surface with ${\geq}$0.97 confidence,
    a common desk-study posture where children's compressed upper body and
    foreshortened limbs diverge markedly from adult training distributions.
    (c)~Rightward lateral trunk lean (0.96 confidence).
    (d)~Correct upright posture (0.91 confidence),
    included to show that the pipeline correctly identifies proper sitting
    as non-deviant rather than triggering false alerts.
    The baseline provides keypoint estimation only and cannot classify posture;
    our pipeline enables category-level posture alerts with high confidence.%
  }
  \label{fig:qualitative}
\end{figure*}

Manual inspection of predicted keypoints on the real-child test set reveals that the COCO-pretrained baseline systematically misplaces head and shoulder keypoints on child subjects, reflecting the anthropometric mismatch between adult training proportions and children's larger head-to-body ratio and shorter limbs.
Wrist and elbow localization is also noticeably degraded, particularly in poses involving desk interaction where children's shorter arms produce joint configurations rarely seen in adult benchmarks.

Figure~\ref{fig:qualitative} illustrates these differences on four representative test images.
Our \ours model produces visibly tighter keypoint clusters across all posture categories, with the most pronounced improvements on the upper-body joints most diagnostic for posture classification: shoulders, elbows, wrists, and head.
In particular, for \textit{desk leaning} and \textit{lateral lean}---postures where the child's compressed upper body deviates most from adult proportions---our model achieves substantially more accurate localization, consistent with the quantitative AP gains reported in Table~\ref{tab:main}.
Beyond keypoint quality, the full pipeline additionally produces per-frame posture classification with confidence scores (visible as bounding boxes and class labels in Figure~\ref{fig:qualitative}), enabling the real-time corrective alerts that motivate this work.

\paragraph{Observed Failure Modes.}
Three recurring failure patterns merit discussion.
First, \textit{self-occlusion under extreme forward lean}: when the child's torso collapses onto the desk, the arms are largely hidden, and wrist keypoints collapse to the shoulder region.
Second, \textit{ambiguous lateral lean magnitude}: subtle lateral shifts ($<5^{\circ}$) are frequently misclassified as correct posture, explaining the relatively lower RR for this category (Table~\ref{tab:e2e_rr}).
Third, \textit{unusual clothing or accessories} (e.g., bulky hooded sweatshirts, wide scarves) occasionally distort the perceived shoulder line, leading to shoulder-keypoint displacement of 10--15 pixels.
These failure cases are consistent with limitations inherent to 2D monocular pose estimation and suggest that depth sensing or temporal modeling could provide meaningful improvements.

\section{Discussion}
\label{sec:discussion}

\subsection{Why AIGC Synthesis Bridges the Domain Gap}
\label{sec:discussion:why}

The +12.5~AP improvement over the COCO-pretrained adult baseline (Table~\ref{tab:main}) stems from two complementary factors.
First, AIGC-synthesized images inherit photographic priors from the generative model's billion-scale training corpus---natural skin tones, wrinkled fabrics, complex backgrounds---while embedding child-specific body proportions, placing them on a visual manifold much closer to real child photographs than either adult training data or CG renders (whose characteristic artifacts such as uniform shading and aliased edges are well-documented sources of shortcut learning~\cite{tobin2017domain, nikolenko2021synthetic}).
Second, combinatorial prompt engineering (Eq.~\ref{eq:prompt}) introduces fine-grained appearance diversity (hairstyles, clothing patterns, backgrounds) tailored to the child domain that cannot be obtained from existing adult-focused datasets.

\subsection{Privacy and Ethical Considerations}
\label{sec:discussion:ethics}

The entire training pipeline requires zero real photographs of minors: Stage~1 uses a parametric body model and Stage~2 generates fictional appearances.
The evaluation test set (Section~\ref{sec:exp:setup}) was collected with informed consent from all participants' guardians and is stored locally without public release.
At deployment, inference runs entirely on-device (\textbf{no video frames leave the device}), with only aggregated posture statistics optionally transmitted.
We use the AIGC-generated images exclusively as intermediate training data and advocate clear data governance protocols for synthetic child imagery.

\subsection{Limitations and Future Work}
\label{sec:discussion:limitations}

\paragraph{Small test set.}
Although our test set (${\sim}$300 images, 4 child subjects, 4 environments) is more diverse than single-subject evaluations, it remains limited in scale. Future work should evaluate on a larger, multi-site benchmark with a broader age range.

\paragraph{Single-scene specialization.}
The pipeline targets desk-study scenarios; extending to other contexts (\eg, piano practice, classrooms) requires new pose templates and prompts, though the architecture generalizes.

\paragraph{Lateral leaning.}
This category achieves only 66.7\% RR (Table~\ref{tab:e2e_rr}), likely due to the subtlety of small lateral shifts and partial self-occlusion. Depth sensing or temporal tracking could help.
Future directions include temporal modeling, cross-domain transfer to other privacy-sensitive applications, and active learning with minimal real data.

\section{Conclusion}
\label{sec:conclusion}

We have presented \ours, an AIGC-based synthetic data pipeline for privacy-preserving child posture estimation.
By decoupling geometric ground truth (from a programmable 3D SMPL-X child model) from photorealistic appearance (from FLUX-1 conditioned via dual ControlNet), our pipeline generates 11{,}900 annotated training images spanning 10 posture categories without requiring a single real child photograph for training.

A custom \texttt{PoseInjectorNode} ensures low annotation drift between the ControlNet conditioning signal and the stored keypoint labels---a design that is critical for training-data integrity and, to our knowledge, not standard in prior synthetic data workflows.
Automated ViTPose-based confidence filtering removes generation failures, and targeted augmentation strategies model the visual degradations of real edge-device camera feeds.

Through direct fine-tuning of RTMPose-M on our synthetic data, the FP16 model achieves 71.2~AP on a real-child test set, a +12.5~AP improvement over the COCO-pretrained adult-data baseline.
After INT8 quantization, the model retains 70.4~AP while running at 22~FPS on a 0.8-TOPS Rockchip RK3568 NPU.
In a single-subject controlled comparison with a commercial posture corrector, our system achieves substantially higher recognition rates and responds ${\sim}$1.8$\times$ faster on average.

These results demonstrate that carefully designed AIGC pipelines can substantially reduce the dependence on real data collection in privacy-sensitive domains.
We hope that \ours serves as a practical blueprint for building ethical, high-performance perception systems where data scarcity and privacy constraints would otherwise be prohibitive.

\section*{Acknowledgments}
We thank members of the open-source developer community for technical consultation on edge-NPU deployment and model quantization.

\section*{Data and Code Availability}
Upon publication, we will release:
(i)~the ComfyUI \texttt{PoseInjectorNode} source code and Blender pose generation plugin;
(ii)~generation configurations including prompt vocabularies, ControlNet adapter versions and weights, sampling parameters, and quality-filtering thresholds;
(iii)~training configuration files and evaluation scripts.
The synthetic training set (11{,}900 images with COCO-format annotations) will be released under a research-use license with usage restrictions on photorealistic child imagery.
The real-child test set cannot be shared due to privacy constraints; anonymized skeleton-only annotations (without images) may be provided upon request.
Note that FLUX-1~Dev is released under a non-commercial license; users intending to deploy the synthesis pipeline commercially should substitute an equivalently capable model with appropriate licensing or obtain a commercial license from the model provider.
The SMPL-X body model is subject to its own non-commercial research license from the Max Planck Society.

\begingroup
\def\bibfont{\hbadness=10000 }
\bibliographystyle{plainnat}
\bibliography{references}
\endgroup



\end{document}